\documentclass[10pt,twocolumn,letterpaper]{article}

\usepackage{cvpr}
\usepackage{times}
\usepackage{epsfig}
\usepackage{graphicx}
\usepackage{amsmath}
\usepackage{amssymb}
\usepackage{algorithm,algorithmic}

\usepackage{subcaption}

\usepackage{mathtools}

\usepackage{array}
\usepackage[pagebackref=true,breaklinks=true,letterpaper=true,colorlinks,bookmarks=false]{hyperref}

\newcolumntype{C}[1]{>{\centering\let\newline\\\arraybackslash\hspace{0pt}}m{#1}}
\cvprfinalcopy % *** Uncomment this line for the final submission

\ifcvprfinal\fi
\begin{document}

\newcommand*{\affmark}[1][*]{\textsuperscript{#1}}
%%%%%%%%% TITLE
\title{EIGEN: Ecologically-Inspired GENetic Approach for Neural Network Structure Searching from Scratch}

\author{
Jian Ren\affmark[1], Zhe Li\affmark[2], Jianchao Yang\affmark[3], Ning Xu\affmark[4], Tianbao Yang\affmark[2], David J. Foran\affmark[1]
\\
{\affmark[1]Rutgers University \quad \affmark[2]The University of Iowa \quad \affmark[3]Bytedance AI Lab \quad \affmark[4]Amazon Go}}

% \and

\maketitle
%\thispagestyle{empty}

%%%%%%%%% ABSTRACT
\begin{abstract}
 Designing the structure of neural networks is considered one of the most challenging tasks in deep learning, especially when there is few prior knowledge about the task domain. In this paper, we propose an Ecologically-Inspired GENetic (EIGEN) approach that uses the concept of succession, extinction, mimicry, and gene duplication to search  neural network structure from scratch with poorly initialized simple network and  few constraints forced during the evolution, as we assume no prior knowledge about the task domain. Specifically, we first use primary succession to rapidly evolve a population of poorly initialized neural network structures into a more diverse population, followed by a secondary succession stage for fine-grained searching based on the networks from the primary succession. Extinction is applied in both stages to reduce computational cost. Mimicry is employed during the entire evolution process to help the inferior networks imitate the behavior of a superior network and gene duplication is utilized to duplicate the learned blocks of novel structures, both of which help to find better network structures. Experimental results show that our proposed approach can achieve similar or better performance compared to the existing genetic approaches with dramatically reduced computation cost. For example, the network discovered by our approach on CIFAR-100 dataset achieves 78.1\%  test accuracy under 120  GPU  hours, compared to 77.0\%  test accuracy in more than 65, 536  GPU hours in~\cite{real2017large}.   

\end{abstract}

%%%%%%%%% BODY TEXT
\section{Introduction}
Deep Convolutional Neural Networks (CNN) have achieved tremendous success among many computer vision tasks~\cite{girshick2015fast,krizhevsky2012imagenet,simonyan2014very}. However, a hand-crafted network structure tailored to one task may perform poorly on another task. Therefore,  it usually requires extensive amount of  human efforts to design an appropriate network structure for a certain task.  

Recently, there are emerging research works ~\cite{assunccao2018using,baker2016designing,bergstra2013making,istrate2018tapas,mendoza2016towards,zoph2017learning} on automatically searching neural network structures for image recognition tasks. In this paper, we focus on optimizing the evolution-based algorithms~\cite{liu2017hierarchical,miikkulainen2017evolving,stanley2002evolving,xie2017genetic} for searching networks from scratch with poorly-initialized networks, such as a network with one global pooling layer, and with few constraints forced during the evolution~\cite{real2017large} as we assume no prior knowledge about the task domain. Existing work along this line of research suffers from either prohibitive computational cost or unsatisfied performance compared with hand-crafted network structures. In~\cite{real2017large}, it costs more than $256$ hours on $250$ GPU for searching neural network structures, which is not affordable for general users. In~\cite{xie2017genetic}, the final learned network structure by their genetic approach achieves about $77\%$ test accuracy on CIFAR-10, even though better performance as 92.9\% could be obtained after fine-tuning certain parameters and modifying some structures on the discovered network. In~\cite{li2018evoltuion}, they firstly aim to achieve better performance with the reduced computational cost by the proposed aggressive selection strategy in genetic approach and more mutations operation to increase diversity which is decreased by the proposed selection strategy. In their work, they reduce computational cost dramatically from more than $64,000$ GPU hours (GPUH) to few hundreds GPUH. However, their approach still suffers performance sacrifice, for example, $90.5\%$ test accuracy compared to $94.6\%$ test accuracy from~\cite{real2017large} on CIFAR-10 dataset.

Inspired by a few key concepts in ecological system, in this paper, we try to improve the genetic approach to achieve better test performance compared to~\cite{real2017large} or competitive performance to hand-crafted network structures~\cite{he2016deep} under limited computation cost~\cite{li2018evoltuion}, but without  utilizing pre-designed architectures~\cite{liu2017progressive,liu2017hierarchical,liu2018darts,zoph2017learning}. Inspired by primary, secondary succession from ecological system~\cite{sahney2008recovery}, we enforce a poorly initialized population of neural network structures to rapidly evolve to a population containing network structures with dramatically improved performance. After the first stage of primary succession, we perform fine-grained search for better networks in a population during the secondary succession stage. During the succession stages, we also introduce an accelerated extinction algorithm to improve the search efficiency. In our approach, we apply the mimicry~\cite{greeney2012feeding} concept to help inferior networks learn the behavior from superior networks to obtain the better performance. In addition, we also introduce the gene duplication to further utilize the novel block of layers that appear in the discovered network structure. 

The contribution of this paper is four-fold and can be summarized as follows:
\begin{itemize}
\item
We proposed an efficient genetic approach to search neural network structure from scratch with poorly initialized network and without limiting the searching space.
Our approach can greatly reduce the computation cost compared to other genetic approaches, where neural network structures are searched from scratch. This is different from some recent works~\cite{elsken2017simple,liu2018darts,pham2018efficient} that significantly restricts the search space.
\item 
We incorporate primary and secondary succession concepts from ecological system into our genetic framework to search for optimal network structures under limited computation cost.
\item We explore the mimicry  concept from ecological system to help search better networks during the evolution and use the gene duplication concept to utilize the discovered beneficial structures.
\item Experimental results show that the  obtained neural network structures achieves better performance compared with existing genetic-based approaches and competitive performance with the hand-crafted network structures.  
\end{itemize}

\section{Related Work}

% \subsection{Comparison}
There is growing interest on automatic searching of neural network architectures from scratch. Methods based on reinforcement learning (RL) show promising results on obtaining the networks with the performance similar or better than human designed architectures~\cite{baker2016designing,zhong2018practical,zoph2016neural}. Zoph \textit{et al}. propose to searching in cells, including a normal cell and a reduction cell, where the final architecture is based on stacking the cells~\cite{zoph2017learning}. The idea of cell based searching is widely adopted in many studies~\cite{cai2018efficient,cai2018path,liu2017progressive,liu2018darts,pham2018efficient,zhong2018blockqnn}. In order to reduce high computational cost, efforts have been done  to avoid training all networks during the searching process from scratch~\cite{baker2017accelerating,bender2018understanding,brock2017smash,cai2018efficient,domhan2015speeding,elsken2017simple,klein2016learning,zhong2017practical}. However, these works require strict hand-designed constraints to reduce computation cost, and comparison with them are not the focus of this paper.

On the other hand,  there emerges a few studies~\cite{real2017large,suganuma2017genetic,xie2017genetic} targeting on network searching using evolutionary approaches. 
In order to have a fair comparison with the RL and evolutionary based approaches, Real~\textit{et al.}~\cite{real2018regularized} conduct the study where the RL and evolutionary approaches are performed under the same searching space. Experiments show the evolutionary approach converges faster than RL.

Therefore, in this paper we focus on the \textit{genetic-based approaches} for searching optimal neural network structures.  Suganuma \textit{et al.} propose the network searching based on Cartesian genetic programming~\cite{harding2008evolution}. However, a pre-defined grid with the fixed row and column is used as the network has to fit in the grid~\cite{suganuma2017genetic} . The studies that have the searching space similar to us are introduced in ~\cite{li2018evoltuion,real2017large,xie2017genetic}, where the network searching starts from poorly-initialized networks and uses few constraints during the evolution.
Since in this paper we focus on achieving better performance with limited computational cost through a genetic approach, we will highlight the differences between our work with the similar  studies~\cite{li2018evoltuion,real2017large,xie2017genetic} in the following from two aspects: reducing computation cost and improving performance. 

In~\cite{real2017large}, the authors encode each individual network structure as a graph into DNA and define several different mutation operations such as IDENTITY and RESET-WEIGHTS to apply to each \textit{parent} network to generate \textit{children} networks. The essential part of this genetic approach is that they utilize a large amount of computation to search the optimal neural network structures in a huge searching space. Specifically, the entire searching procedure costs more than $256$ hours with $250$ GPUs to achieve $94.6\%$ test accuracy from the learned network structure on CIFAR-10 dataset,  which is not affordable for general users. 

Due to prohibitive computation cost, in~\cite{xie2017genetic} the authors impose restriction on the neural network searching space. In their work, they only learn one block of network structure and stack the learned block by certain times in a designed routine to obtain the best network structure. Through this mechanism, the computation cost is reduced to several hundreds GPU hours, however, the test performance of the obtained network structure is not satisfactory, for example, the found network achieves $77\%$ test accuracy on CIFAR-10, even though fine-tuning parameters and modifying certain structures on the learned network structure could lead to the test accuracy as 92.9\%.
% the better performance.

%Genetic Algorithms for Evolving Deep Neural Networks
In~\cite{li2018evoltuion}, they aim to achieve better performance from  automatically learned network structure with limited computation cost in the course of evolution, which is not brought up previously. Different from restricting the search space to reduce computational cost~\cite{dufourq2017eden,xie2017genetic}, they propose the aggressive selection strategy to eliminate the weak neural network structures in the early stage. However, this aggressive selection strategy may decrease the diversity which is the nature of genetic approach to improve performance. In order to remedy this issue, they define more mutation operations such as add\_fully\_connected or add\_pooling. Finally, they reduce computation cost dramatically to 72 GPUH on CIFAR-10. However, there is still performance loss in their approach. For example, on CIFAR-10 dataset, the test accuracy of the found network is about $4\%$ lower than~\cite{real2017large}. 

At the end of this section, we highlight that our work is in the line of ~\cite{li2018evoltuion}.
Inspired from ecological concepts, we propose the Ecologically-Inspired GENetic approach (EIGEN) for neural network structure search by evolving the networks through rapid succession, and explore the mimicry and gene duplication along the evolution.

\section{Approach}
Our genetic approach for searching the optimal neural network structures follows the standard procedures: i) initialize population in the first generation with simple network structures; ii) evaluate the \textit{fitness} score of each neural network structure (fitness score is the measurement defined by users for their purpose such as validation accuracy, number of parameters in network structure, number of FLOP in inference stage, and so on); iii) apply a selection strategy to decide the surviving network structures based on the fitness scores; iv) apply mutation operations on the survived \textit{parent} network structures to create the \textit{children} networks for next generation. The last three steps are repeated until the convergence of the fitness scores. Note that in our genetic approach, the \textit{individual} is denoted by an acyclic graph with each node representing a certain layer such as \textit{convolution}, \textit{pooling} and \textit{concatenation} layer. A \textit{children} network can be generated from a \textit{parent} network through a mutation procedure. A \textit{population} includes a fixed number of networks in each generation, which is set as 10 in our experiments. For details of using genetic approach to search neural network structures, we refer the readers to~\cite{li2018evoltuion}. In the following, we apply the ecological concepts of succession, extinction, mimicry and gene duplication  to the genetic approach for an accelerated search of neural network structures. 

\subsection{Evolution under Rapid Succession}

Our inspiration comes from the fact that in an ecological system, the population is dominated by diversified fast-growing individuals during the primary succession, while in the secondary succession, the population is dominated by more competitive individuals~\cite{sahney2008recovery}. 
% Therefore, we treat all the networks during each generation of the evolution process as a \textit{community} and treat the individual networks in each generation as a population, instead of focusing on evolving a single network~\cite{real2017large}. 
Therefore, we treat all the networks during each generation of the evolution process as a \textit{population} and focus on evolving the population instead of on a single network~\cite{real2017large}. 

With this treatment, we propose a two-stage \textit{rapid succession} for accelerated evolution, analogous to the ecological succession.
The proposed rapid succession includes a primary succession, where it starts with a community consisting of a group of poorly initialized individuals which only contains one global pooling layer, and a secondary succession which starts after the primary succession. 
In the primary succession, a large search space is explored to allow the community grow at a fast speed, and a relatively small search space is used in the secondary succession for fine-grained search. 

In order to depict how the search space is explored, we define \textit{mutation step-size} $m$ as the maximum mutation iterations between the parent and children. 
The actual mutation step for each child is uniformly chosen from $[1,m]$.
In the primary succession, in order to have diversified fast-growing individuals, a large mutation step-size is used in each generation so the mutated children could be significantly different from each other and from their parent. 
Since we only go through the training procedure after finishing the entire mutation steps, the computation cost for each generation will not increase with the larger step-size. 
In the secondary succession, we adopt a relative small mutation step-size to perform a fine-grained search for network structures.

Each mutation step is randomly selected from the nine following operations including:
\begin{itemize}
\item INSERT-CONVOLUTION: A convolutional layer is randomly inserted into the network. The inserted convolutional layer has a default setting with kernel size as 3$\times$3, number of channels as 32, and stride as 1. The convolutional layer is followed by batch normalization~\cite{ioffe2015batch} and Rectified Linear Units~\cite{krizhevsky2012imagenet}.
\item INSERT-CONCATENATION: A concatenation layer is randomly inserted into the network where two bottom layers share the same size of feature maps.
\item INSERT-POOLING: A pooling layer is randomly inserted into the network with kernel size as 2$\times$2 and stride as 2.
\item REMOVE-CONVOLUTION: The operation randomly remove a convolutional layer.
\item REMOVE-CONCATENATION: The operation randomly remove a concatenation layer.
\item REMOVE-POOLING: The operation randomly remove a pooling layer.
\item ALTER-NUMBER-OF-CHANNELS, ALTER-STRIDE, ALTER-FILTER-SIZE: The three operations modify the hyper-parameters in the convolutional layer. The number of channels is randomly selected from a list of $\left \{  16, 32, 48, 64, 96 \right \}$; the stride is randomly selected from a list of $\left \{  1, 2 \right \}$; and the filter size is randomly selected from $\left \{  1\times1,  3\times3\right \}$.
\end{itemize}
% Thus for the purpose of the accelerated evolution, 

% \subsubsection{Accelerated Extinction}
During the succession, we employ the idea from previous work~\cite{li2018evoltuion} that only the best individual in the previous generation will survive. However, instead of evaluating the population in each generation after all the training iterations, it is more efficient to extinguish the individuals that may possibly fail at early iterations, especially during the primary succession where the diversity in the population leads to erratic performances. Based on the assumption that a better network should have better fitness score at earlier training stages, we design our extinction algorithm as follows. 

To facilitate the presentation, we denote $n$ as the population size in each generation, $T_1$ and $T_2$ as the landmark iterations, $f_{g,i,T_1}$ and $f_{g,i,T_2}$ as fitness scores (validation accuracy used in our work) of the $i^{th}$ network in the $g^{th}$ generation after training $T_1$ and $T_2$ iterations, $v_{g, T_1}$ and $v_{g, T_2}$ as threshold to eliminate weaker networks at $T_1$ and $T_2$ iterations in the $g^{th}$ generation. 
%$p = 5$ and $q = 2$  
In the $g^{th}$ generation, we have fitness scores for all networks $\mathcal{F}_{g,T_1} = \{f_{g,i,T_1}, i = 1, \cdots, n\}$ and $\mathcal{F}_{g,T_2} =\{ f_{g,i,T_2}, i = 1, \cdots, \hat{n}\}$  after training $T_1$ and $T_2$ iterations, respectively. Note that $\hat{n}$ can be less than $n$ since  weaker networks are eliminated after $T_1$ iterations. 
The thresholds $v_{g, T_1}$ and $v_{g, T_2}$ are updated at $g^{th}$ iteration as
\begin{equation}
    v_{g, T_1} = \text{max} \Big(S(\mathcal{F}_{g, T_1})_p,  \hspace{0.1cm} v_{g-1, T_1}\Big)
\label{eq:T1}
\end{equation}
and 
\begin{equation}
    v_{g, T_2} = \text{max} \Big( S(\mathcal{F}_{g, T_2})_q,  \hspace{0.1cm} v_{g-1, T_2}\Big)
\label{eq:T2}
\end{equation}
where $S(.)$ is a sorting operator in decreasing order on a list of values and the subscripts $p$ and $q$ represents $p^{th}$ and $q^{th}$ value after the sorting operation, $p$ and $q$ are the hyper-parameters.

For each generation, we perform the following steps until the convergence of the fitness scores: (i) train the population for $T_1$ iterations, extinguish the individuals with fitness scores less than $ v_{g, T_1}$; (ii) train the remaining population for $T_2$ iterations, and distinguish the population with fitness scores less than $v_{g, T_2}$; (iii) the survived individuals are further trained till convergence and the best one is chosen as the parent for next generation.  
 %The updating of $v_1$ and $v_2$ is depending on $f_1p$ and $f_2q$ in each generation.
The details for the extinction algorithm are described in Algorithm~\ref{alg:aae}.

\begin{algorithm}[t]
\caption{Algorithm for Extinction}
\label{alg:aae}
\begin{algorithmic}[1]
% \small \STATE{}
    %\SetKwInOut{Input}{Input}
   \STATE \textbf{Input:} $T_1$, $T_2$, $v_{0,T_1}$, $v_{0,T_2}$, $p$, $q$
    \FOR{$g = 1 \cdots, G$}
    \STATE Obtain $\mathcal{F}_{g, T_1} = \{ f_{g,i,T_1},i=1, ...,n\}, n=10$ by training all individuals for $T_1$ iterations\\
    \STATE Update $v_{g, T_1}$ based on Eq.~\ref{eq:T1} \\
    \STATE Extinguish the individuals with fitness value less than $v_{g, T_1}$\\
    \STATE Obtain $\mathcal{F}_{g, T_2} = \{ f_{g,i,T_1},i=1, ...,\hat{n}\}$ by training the remain individuals for $T_2$ iterations\\
    \STATE Update $v_{g, T_2}$ based on Eq.~\ref{eq:T2} \\
    \STATE Extinguish the individuals with fitness value less than $v_{g, T_2}$\\
    \STATE Train the remain individuals for $T_3$ iterations and select the best one as parent
    \ENDFOR
\end{algorithmic}
\end{algorithm}

%--------------------------------Figure: duplication--------------------------------------------
\begin{figure}[t]
\begin{center}
\includegraphics[width=1\columnwidth]{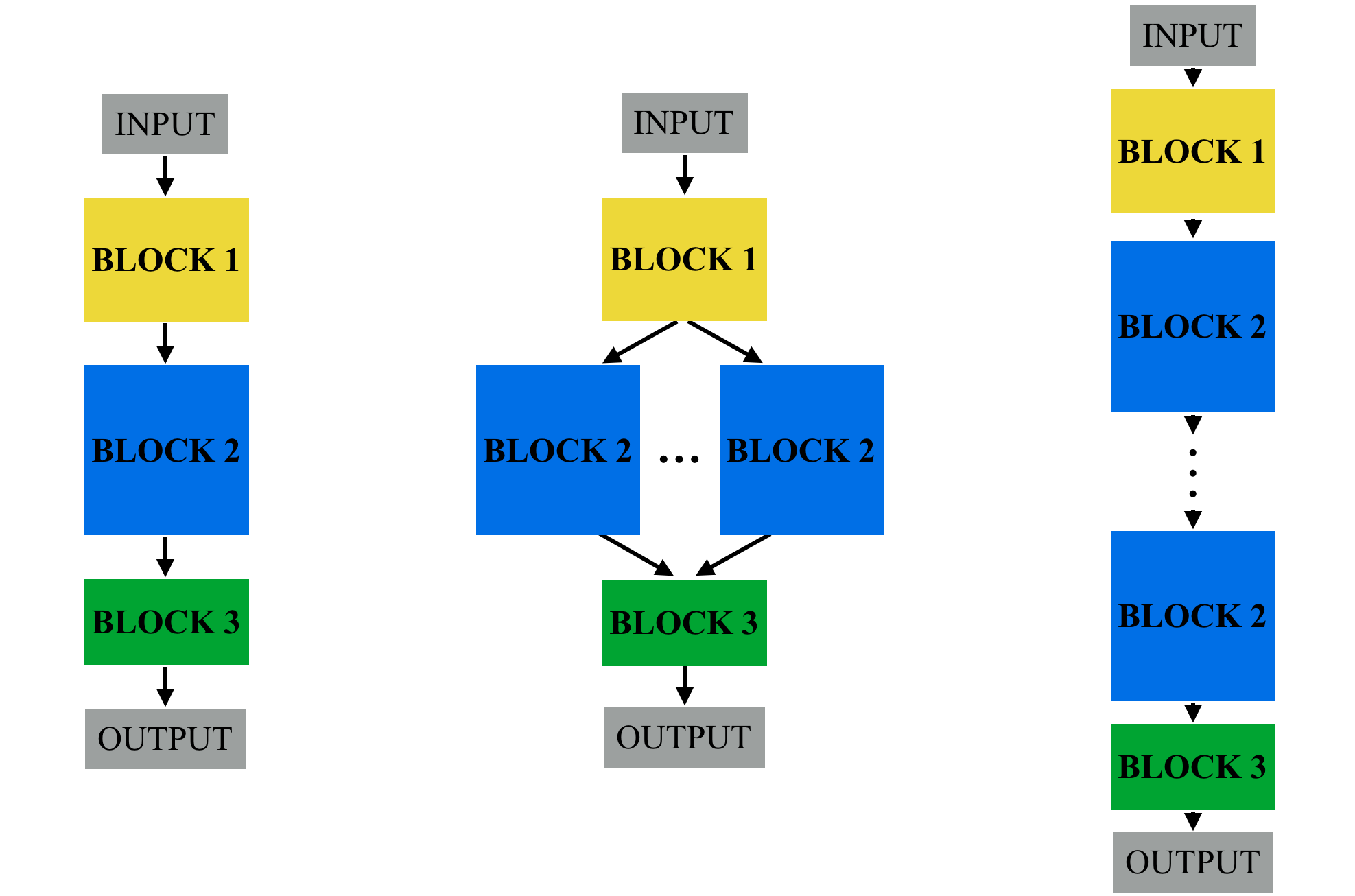}
\caption{Example of duplication. The image on the left shows the structure discovered after the rapid succession, where each block includes a number of layers with the same size of feature maps. The image in the middle and right are two examples of the duplication that the Block 2 undergoes different combination to create new architectures.}
\label{duplication_exp}
\end{center}
\end{figure}
%--------------------------------Figure: duplication--------------------------------------------

%---------------------------Table Cifar10, Cifar100------------------------------
\begin{table*}
\begin{center}
\begin{tabular}{C{5cm} | C{2cm}  C{2cm} C{2cm} C{3cm} }
\hline
Model & PARAMS. & C10+  & C100+ & Comp Cost   \\ \hline\hline
MAXOUT~\cite{goodfellow2013maxout} & - &90.7\% & 61.4\% & - \\ \hline
Network In Network~\cite{lin2013network} & - & 91.2\% & 64.3\%& - \\ \hline
ALL-CNN~\cite{springenberg2014striving} & 1.3 M & 92.8\% & 66.3\% & -\\ \hline
DEEPLY SUPERVISED~\cite{lee2015deeply} & - & 92.0\% & 65.4\% & -\\ \hline
HIGHWAY~\cite{srivastava2015highway} & 2.3 M & 92.3\% & 67.6\% & - \\ \hline
RESNET~\cite{he2016deep}& 1.7 M & 93.4\% & 72.8\%& - \\ \hline
DENSENET ($k = 40, l = 100$)~\cite{huang2017densely} & 25.6 M & 96.5\% & 82.8\%& - \\ \hline
Teacher Network & 17.2 M & 96.0\% & 82.0\% & - \\\hline\hline
EDEN~\cite{dufourq2017eden}&0.2 M&74.5\% &- &- \\ \hline
Genetic CNN~\cite{xie2017genetic} & - & 92.9\% & 71.0\% & 408 GPUH \\ \hline
LS-Evolution~\cite{real2017large}& 5.4 M& 94.6\% &- & 64,000 GPUH \\ 
LS-Evolution~\cite{real2017large}& 40.4 M& -& 77.0\% & $>$ 65,536 GPUH \\ \hline
AG-Evolution~\cite{li2018evoltuion} & - & 90.5\%&- & 72 GPUH\\ 
AG-Evolution~\cite{li2018evoltuion} & - & -& 66.9\%& 136 GPUH\\ \hline\hline
EIGEN & 2.6 M & \textbf{94.6}\% & - & 48 GPUH\\ \hline
EIGEN & 11.8 M & - &  \textbf{78.1}\% &  120 GPUH\\ \hline
\end{tabular}
\end{center}
\caption{Comparison with hand-designed architectures and automatically discovered architectures using genetic algorithms. The C10+ and C100+ columns indicate the test accuracy achieved on data-augmented CIFAR-10 and CIFAR-100 datasets, respectively. The PARAMS. column indicates the number of parameters in the discovered network.}
\label{compare_resutls}
\end{table*}
%---------------------------Table Cifar10, Cifar 100-------------------------------

%imitation learning
% \subsection{Knowledge Distillation}
\subsection{Mimicry}
%Inspired by adaptation~\cite{orr2005genetic} 
%fitness~\cite{romero2014fitnets, yim2017gift}
%Many insects have a larval form that is optimized for extracting energy and nutrients from the envi-ronment and a completely different adult form that is optimized for the very different requirementsof traveling and reproduction.

% =========NOTE: mimicry is a method that
% =========NOTE: In our approach, we guide the inferior networks  
In biology evolution, mimicry is a phenomenon that one species learn behaviours from another species. For example, moth caterpillars learn to imitate body movements of a snake so that they could scare off predators that are usually prey items for snakes~\cite{greeney2012feeding}. 
The analogy with mimicry signifies that we could force inferior networks to adopt (learn) the behaviors, such as statistics of feature maps~\cite{romero2014fitnets,yim2017gift} or logits~\cite{bucilu2006model,hinton2015distilling}, from superior networks in designing neural network structure during the evolution.

In our approach, we force the inferior networks to learn the behavior of a superior network by generating similar distribution of logits in the evolution procedure. Since learning the distribution of logits from the superior network gives more freedom for inferior network structure, compared to learning statistics of feature maps. This is in fact the knowledge distillation proposed in~\cite{hinton2015distilling}.  
% Besides the rapid succession, we explore the possibility to incorporate knowledge distillation in the evolution algorithm.
More specifically, for the given training image $\mathbf{x}$ with one-hot class label $\mathbf{y}$, we define $\mathbf{t}$ as the logits predicted from the pre-trained superior network, and $\mathbf{s}$ as the logits predicted by the inferior network. We use the following defined $\mathcal{L}_K$ as the loss function to encode the prediction discrepancy between inferior and superior networks as well as the difference between inferior networks prediction and ground truth annotations during the evolution:
\begin{equation}
\begin{split}
\mathcal{L}_K= (1 - \alpha )\mathcal{L}_C(\mathbf{y}, \mathcal{H}(\mathbf{s})) +  
\alpha T^2\mathcal{L}_C\left (\mathcal{H}\left (\frac{\mathbf{s}}{T} \right ),\mathcal{H} \left (\frac{\mathbf{t}}{T}\right )  \right )
\label{distillation}
\end{split}
\end{equation}
where $\mathcal{H}(.)$ is the \textit{softmax} function, $\mathcal{L}_C$ is the cross-entropy of two input probability vectors such that 
\begin{equation}
\mathcal{L}_C(\mathbf{y}, \mathcal{H}(\mathbf{s})) = -\sum_{k}\mathbf{y}_k \mathrm{log} \mathcal{H}(\mathbf{s}_k) , 
\end{equation}
% $T$ is a temperature parameter 
$\alpha$ is the ratio controlling two loss terms and $T$ is a hyper-parameter.
We adopt the terms from knowledge distillation~\cite{hinton2015distilling} where student network and teacher network represent the inferior network and superior network, respectively.
We fix $T$ as a constant.
While the target of neural network search is to find the optimal architecture, mimicry is particularly useful when we want to find a small network for applications where inference computation cost is limited.

\subsection{Gene Duplication}
During the primary succession, the rapid changing of network architectures leads to the novel beneficial structures decoded in DNA~\cite{real2017large} that are not shown in the previous hand-designed networks. To further leverage the automatically discovered structures, we propose an additional mutation operation named \textit{duplication} to simulate the process of gene duplication since it has been proved as an important mechanism for obtaining new genes and could lead to evolutionary innovation~\cite{zhang2003evolution}. In our implementation, we treat the encoded DNA as a combination of blocks. For each layer with the activation map defined as $N \times D \times W \times H$, where $N, D, W, H$ denote the batch size, depth, width and height, respectively, the block includes the layers with activation map that have the same $W$ and $H$.
% where each block includes the layers with the same size of feature maps. 
As shown in Figure~\ref{duplication_exp}, the optimal structure discovered from the rapid succession could mutate into different networks by combining the blocks in several ways through the duplication. We duplicate the entire block instead of single layer because the block contains the beneficial structures discovered automatically while simple layer copying is already an operation in the succession.

%------------------------------Figure: cifar10 achitecture--------------------------------------
\begin{figure*}
\centering
\begin{subfigure}{1\textwidth}
\centering
  \includegraphics[width=0.95\linewidth]{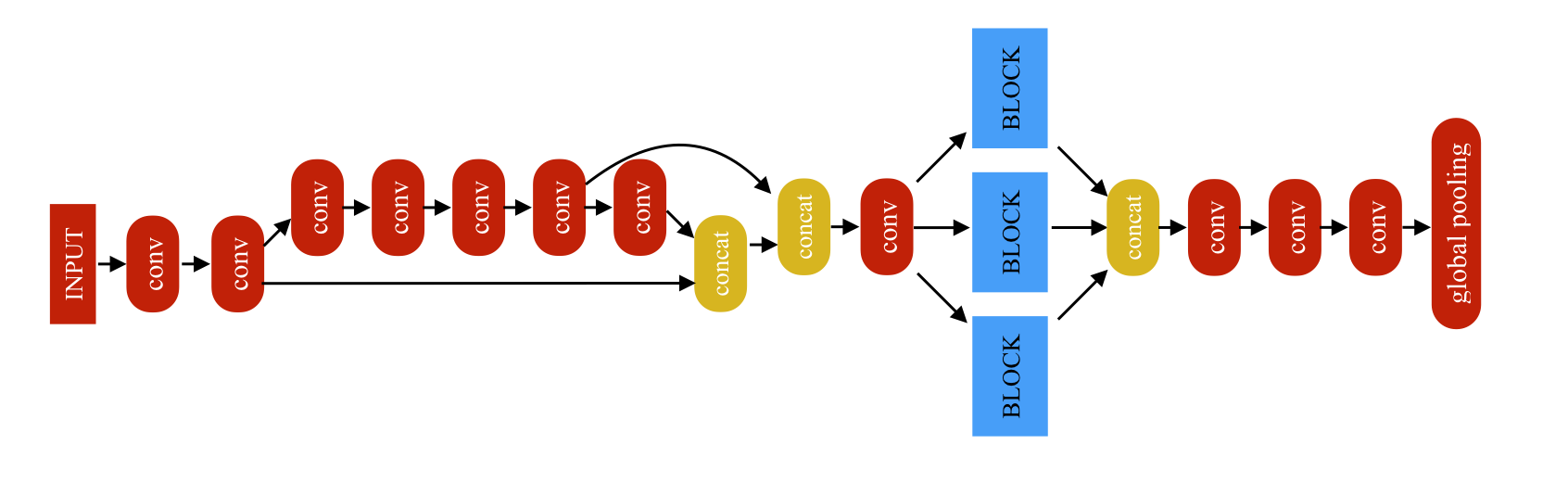}
  \caption{The discovered network architecture using the proposed method on CIFAR-10 dataset that includes convolutional layers, concatenation layers and global pooling layer.}
  \label{cifar_10_network_s}
\end{subfigure} \\
\begin{subfigure}{1\textwidth}
\centering
  \includegraphics[width=0.95\linewidth]{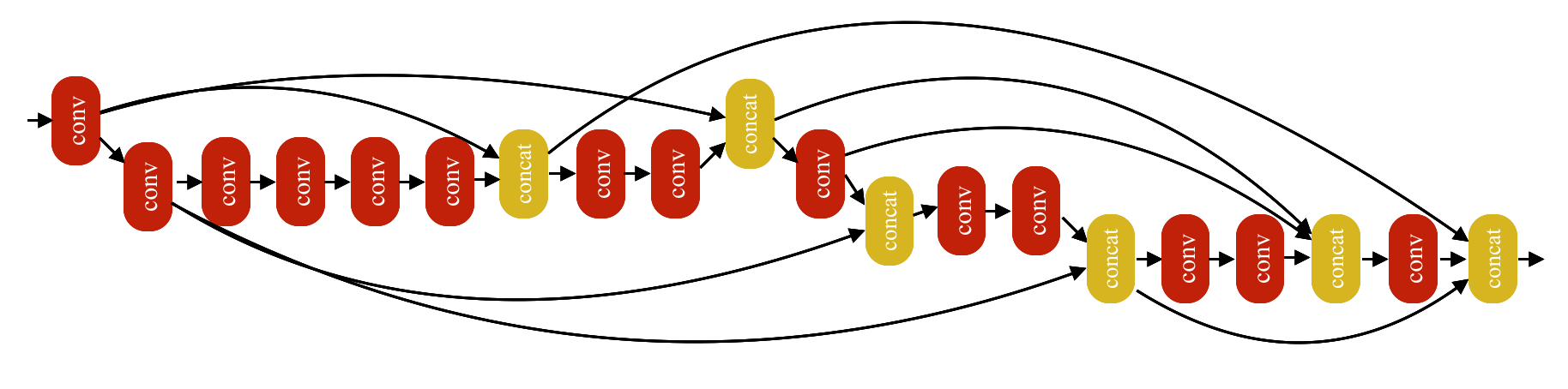}
  \caption{The detailed architecture of the BLOCK shown in (a).}
  \label{cifar_10_network_b}
\end{subfigure}
\caption{Discovered neural network structure for CIFAR-10 dataset.} 
\label{cifar_10_network}
\end{figure*}
%------------------------------Figure: cifar10 achitecture--------------------------------------

%--------------------------------Figure: mutation_exp--------------------------------------------
\begin{figure}
\centering
\begin{subfigure}{0.5\textwidth}
\centering
  \includegraphics[width=1\linewidth]{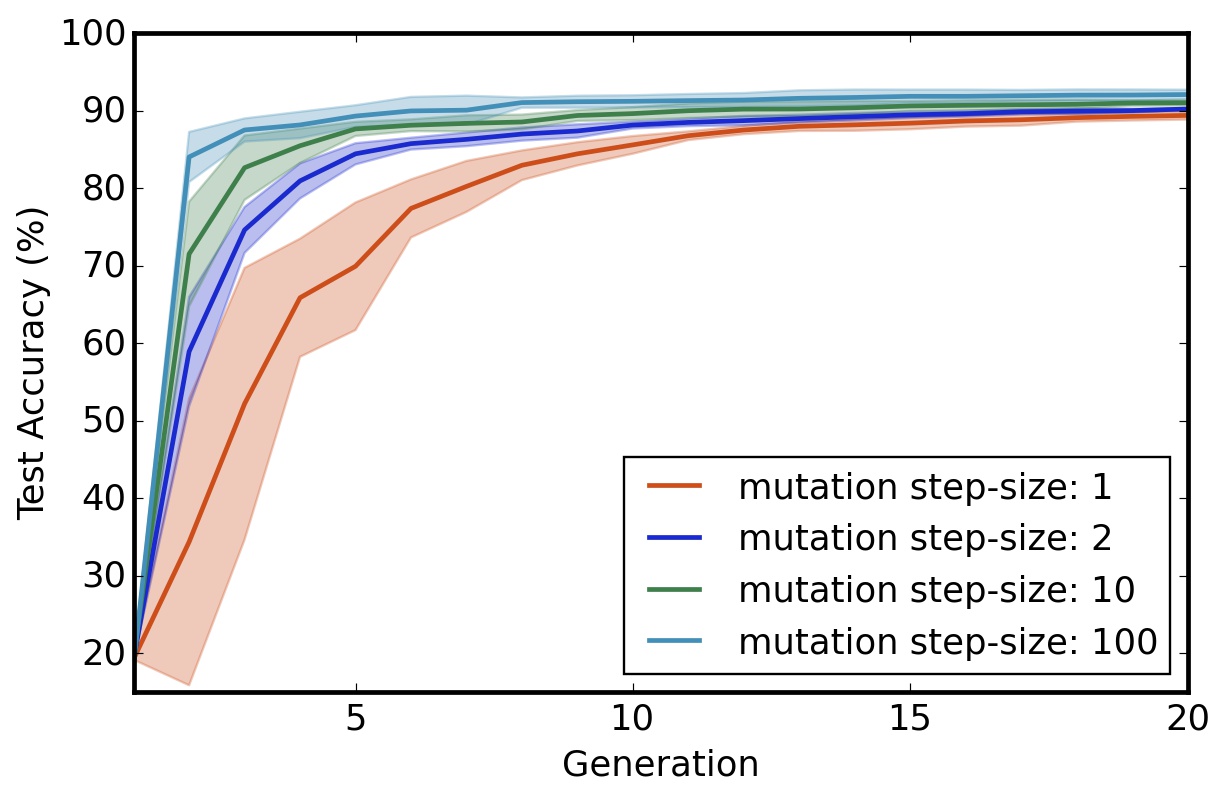}
  \caption{Primary succession using the mutation step-size as 1, 2, 10 and 100.}
  \label{step_size_s}
\end{subfigure}\\
\begin{subfigure}{0.5\textwidth}
\centering
  \includegraphics[width=1\linewidth]{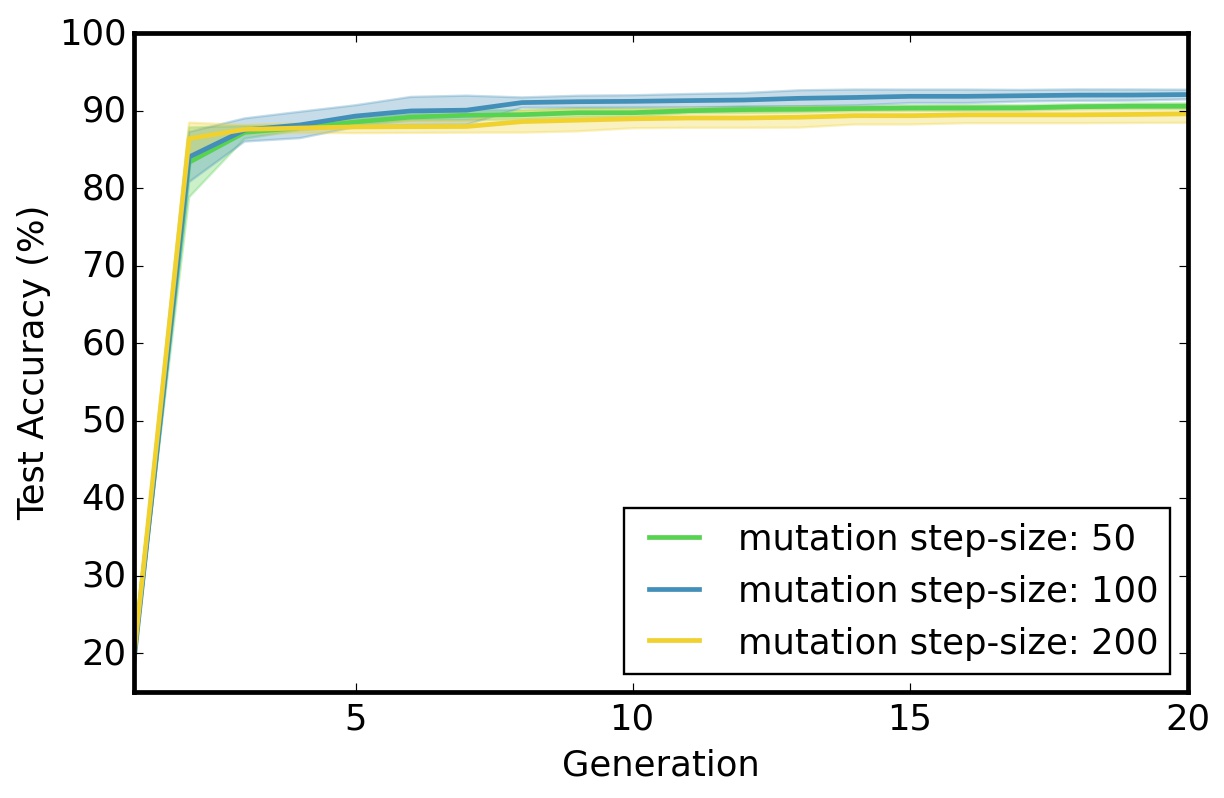}
  \caption{Primary succession using the muation step-size as 50, 100 and 200.}
  \label{step_size_l}
\end{subfigure}
\caption{The effect of different mutation step-size for the primary succession on CIFAR-10. The solid lines show the average test accuracy over five experiments of the individuals with the highest accuracy in each generation. The shaded area around each line has a width of standard deviation $\pm \sigma $. In general, the larger mutation step-size, the faster the convergence of fitness score.} 
\label{mutation_exp}
\end{figure}
%--------------------------------Figure: mutation_exp--------------------------------------------

\section{Experimental Results and Analysis}
In this section, we report the experimental results of using EIGEN for structure search of neural networks. We first describe the experiment setup including datasets preprossessing and training strategy in Subsection~\ref{ss:experiment-setup} and show the comparison results in Subsection~\ref{ss:experiment-results}.
Following that, we analyze the experimental results in Subsection~\ref{ss:rapid-succession}  with regard to each component of our approach.
% and~\ref{ss:kd}, respectively.  
\subsection{Experiment Setup}\label{ss:experiment-setup}
\paragraph{Datasets.}
The experiments are conducted on two benchmark datasets including CIFAR-10~\cite{krizhevsky2009learning} and CIFAR-100~\cite{krizhevsky2009learning}. The CIFAR-10 dataset contains 10 classes with 50, 000 training images and 10, 000 test images. The images have the size of 32$\times$32. The data augmentation is applied by a Global Contrast Normalization (GCN) and ZCA whitening~\cite{goodfellow2013maxout}. The CIFAR-100 dataset is similar to CIFAR-10 except it includes 100 classes.

\paragraph{Training Strategy and Details.}
During the training process, we use mini-batch Stochastic Gradient Descent (SGD) to train each individual network with the batch size as 128, momentum as 0.9, and weight-decay as 0.0005. Each network is trained for a maximum of 25, 000 iterations. The initial learning rate is 0.1 and is set as 0.01 and 0.001 at 15, 000 iterations and 20, 000 iterations, respectively. The parameters in Algorithm~\ref{alg:aae} are set to $T_1 = 5, 000$, $T_2 = 15, 000$, $T_3 = 5, 000$, $p = 5$, and $q = 2$.
For the mimicry, we set $T$ to 5 and $\alpha$ to 0.9 in Eq.~\ref{distillation}. 
The teacher network is an ensemble of four Wide-DenseNet ($k=60, l=40$)~\cite{huang2017densely}. The fitness score is validation accuracy from validation set. The primary succession ends when the fitness score saturates and then the secondary succession starts. The entire evolution procedure is terminated when the fitness score converges. 
% We use the teacher network that has the test accuracy as 96.0\% on CIFAR-10 and 82.0\% on CIFAR-100. 
Training is conducted with \textit{TensorFlow}~\cite{abadi2016tensorflow}.

We directly adopt the hyper-parameters developed on CIFAR-10 dataset to CIFAR-100 dataset.  
%The EIGEN is developed on CIFAR-10 and applied to CIFAR-100 without any changes. 
The experiments are run on a machine that has one Intel Xeon  E5-2680 v4 2.40GHz CPU and one Nvidia Tesla P100 GPU.  %measure the computational cost. 

\subsection{Comparison Results}\label{ss:experiment-results}

The experimental results shown in Table~\ref{compare_resutls} justify the proposed approach are competitive with hand designed networks. Compared with the evolution-based algorithms, we can achieve the best results with the minimum computational cost. For example, we obtain similar results on the two benchmark datasets compared to ~\cite{real2017large}, but our approach is 1,000 times faster. 
Also, the number of parameters of the networks found by our approach on the two datasets are more than two times smaller than LS-Evolution~\cite{real2017large}. 

We show the discovered network architecture using our proposed method on CIFAR-10 dataset in Figure~\ref{cifar_10_network}, where Figure~\ref{cifar_10_network_s} shows the engire network and Figure~\ref{cifar_10_network_b} represents the detailed architecture in the BLOCK of Figure~\ref{cifar_10_network_s}.
% More details for the network architectures are reported in the supplementary materials.

%\section{Analysis}
\subsection{Analysis}\label{ss:rapid-succession}
\paragraph{Effect of Primary Succession.}
We show the results on different mutation step-size for the primary succession in Figure~\ref{mutation_exp}. The solid lines show average test accuracy of the best networks among five experiments and the shaded area represents the standard deviation $\sigma$ in each generation among five experiments. Larger mutation step-size, such as 100,  leads to the faster convergence of fitness score compared with the smaller mutation step-size, as shown in Figure~\ref{step_size_s}. However, no further improvement is observed by using too large mutation step-size, such as 200, as shown in Figure~\ref{step_size_l}. 
% The difference may resulted from the kinds of mutation operations defined in the succession.
% Note, reconsider 

\begin{figure}
\centering
\begin{subfigure}{0.5\textwidth}
\centering
  \includegraphics[width=1\linewidth]{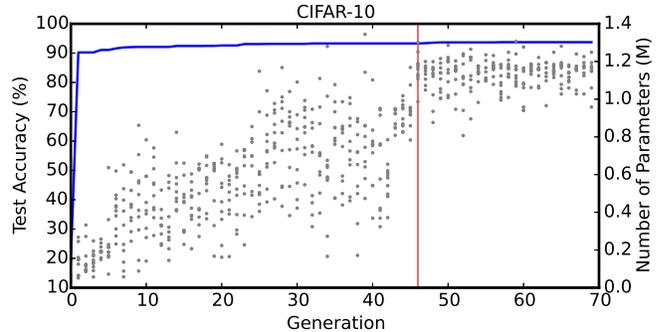}
  \caption{Experiment on CIFAR-10 dataset.}
  \label{cifar_10}
\end{subfigure}\\
\begin{subfigure}{0.5\textwidth}
\centering
  \includegraphics[width=1\linewidth]{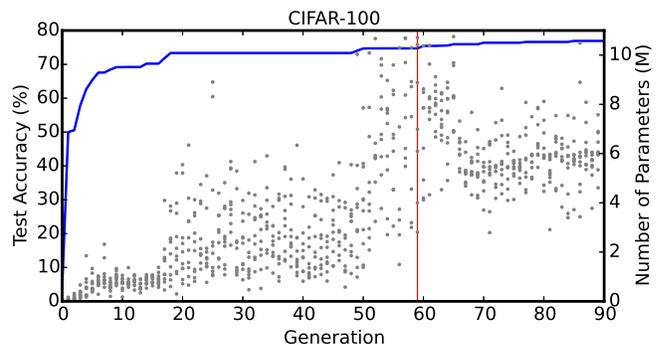}
  \caption{Experiment on CIFAR-100 dataset.}
  \label{cifar_100}
\end{subfigure}
\caption{The progress of rapid succession on CIFAR-10 (a) and CIFAR-100 (b). The blue line is the test performance of the best individual in each generation. The gray dots show the number of parameters of the individuals in each generation. The red line denotes the generation where primary succession ends.} 
\label{two_datasets}
\end{figure}

\paragraph{Effect of Secondary Succession.}
We further analyze the effect of the secondary succession during the evolution process. After the primary succession, we utilize the secondary succession to search the networks with a smaller searching space. We adopt small mutation step-size for the purpose of fine-grained searching based on the survived network from previous generation. Figure~\ref{two_datasets} shows the example evolution on CIFAR-10 and CIFAR-100 during the rapid succession. We use mutation step-size 100 and 10 for primary succession and secondary succession, respectively.
The blue line in the plots shows performance of the best individual in each generation. The gray dots show the number of parameters for the population in each generation, and the red line indicates where the primary succession ends. 
The accuracy on the two datasets for the secondary succession shown in Table~\ref{secondary_resutls} demonstrates that small mutation step-size is helpful for searching better architectures in the rapid succession.

%--------------------------------Figure: cifar10 & cifar100 size acc-------------------------------------------
%------------------------------Figure: cifar10 & cifar100 size acc---------------------------------------

\paragraph{Analysis on Mimicry.}\label{ss:kd}
In order to analyze the effect of mimicry, we consider the situation where only primary and secondary succession are applied during the evolution. Both the duplication and mimicry are disabled.  We denote the method as \textbf{EIGEN w/o mimicry and duplication}. We compare  \textbf{EIGEN w/o mimicry and duplication} with the approach where mimicry is enabled and denote it as \textbf{EIGEN w/o duplication}. The comparison between \textbf{EIGEN w/o mimicry and duplication} and \textbf{EIGEN w/o duplication} in Table~\ref{distill_resutls} proves the effectiveness of the mimicry during the rapid succession.

\begin{table}
      	\centering
        \begin{tabular}{C{3.2cm} | C{0.96cm} C{0.96cm}}
        \hline
        Succession&C10+&C100+  \\ \hline\hline
        Primary Succession&93.3\%&74.7\% \\ \hline
        Secondary Succession&93.7\%&76.9\% \\ \hline
        \end{tabular}
        \caption{The results of secondary succession during the evolution. After the primary succession, the smaller mutation step-size is adopted to search the better network architectures. The accuracy on both CIFAR-10 and CIFAR-100 are improved.}
        \label{secondary_resutls}
 \end{table}
 
 \begin{table}
      \centering
        \begin{tabular}{C{3cm} | C{1cm}  C{1cm}}
        \hline
        Method & C10+  & C100+  \\ \hline\hline
        EIGEN w/o mimicry and duplication&  92.4\%  & 74.8\%\\ \hline
        EIGEN w/o duplication&  93.7\% & 76.9\% \\ \hline
        \end{tabular}
        \caption{Analysis of mimicry during the raipd succession.}
        \label{distill_resutls}
\end{table}

\paragraph{Effect of Gene Duplication.} After the rapid succession, the duplication operation is applied to leverage the automatically discovered structures. To analyze the effect of gene duplication, we denote the approach without duplication as \textbf{EIGEN w/o duplication} and show the results on CIFAR-10 and CIFAR-100 in Table~\ref{duplication_resutls}. Although more parameters are induced in the networks by duplication, the beneficial structures contained in the block can actually contribute to the network performance through duplication.

%---------------------------Table duplication Cifar10, Cifar100------------------------------
\begin{table}[h]
\begin{center}
\centerline{
\begin{tabular}{C{3.33cm} | C{2.1cm} C{2.3cm} }
\hline
Method & C10+ (PARAMS.) & C100+ (PARAMS.)  \\ \hline\hline
EIGEN w/o duplication & 93.7\% (1.2 M) & 76.9\% (6.1 M) \\ 
EIGEN & 94.6\% (2.6 M) &  78.1\% (11.8 M) \\ \hline
\end{tabular}
}
\end{center}
\caption{Analysis of the gene duplication operation on CIFAR-10 and CIFAR-100. The performance on the two datasets is improved with more parameters on the networks discovered from gene duplication.}
\label{duplication_resutls}
\end{table}
%----------------------------Table duplication Cifar10, Cifar100----------------------------------

Furthermore, we analyze the effect of mimicry on the network after the gene duplication. We denote the best network found by our approach as \textbf{EIGEN network}. By utilizing the mimicry to train the network from scratch, which is \textbf{EIGEN network w mimicry}, the networks obtain the improvement as 1.3\% and 4.2\% on CIFAR-10 and CIFAR-100, respectively, compared with the network trained from scratch without mimicry, which is \textbf{EIGEN network w/o mimicry}.

%---------------------------Table distillation loss Cifar10, Cifar100------------------------------
\begin{table}[h]
\begin{center}
\centerline{ 
\begin{tabular}{C{4.4cm} | C{1.5cm}  C{1.5cm}}
\hline
Method & C10+  & C100+  \\ \hline\hline
% Succession w distillation w/o distill & 92.6\%  \\ \hline\hline
EIGEN network w/o mimicry & 93.3\% &73.9\% \\
EIGEN network  w mimicry & 94.6\% & 78.1\%\\  \hline
\end{tabular}
}
\end{center}
\caption{Analysis of mimicry after the gene duplication. }
\label{distill_resutls_2}
\end{table}
%----------------------------Table distillation loss Cifar10, Cifar100----------------------------------

\section{Discussion and Conclusions}
% limitation ...

In this paper, we propose an Ecologically-Inspired GENetic Approach (EIGEN) for searching neural network architectures automatically from scratch, with poor initialization networks, such as a network with one global pooling layer, and few constraints forced during the searching process.  Our searching space follows the work in~\cite{li2018evoltuion,real2017large} and we introduce rapid succession, mimicry and gene duplication in our apporach to make the search more efficient and effective. The rapid succession and mimicry could evolve a population of networks into an optimal status under the limited computational resources. With the help of gene duplication, the performance of the found network could be boosted without sacrificing any computational cost. The experimental results show the proposed approach can achieve competitive results on CIFAR-10 and CIFAR-100 under dramatically reduced computational cost compared with other genetic-based algorithms.

Admittedly, compared with other searching neural network algorithms~\cite{liu2018darts,pham2018efficient} which aim to searching network under limited computation resource, our work has the slightly higher error rate. But our genetic algorithm requires little prior domain knowledge from human experts, and is more ``complete-automatic'' compared with other semi-automatic searching neural network approaches~\cite{liu2018darts,pham2018efficient}, which require more advanced initialization, carefully designed cell-based structures and much more training iterations after the searching process.  
Such comparison, although unfair, still indicates that more exploration is needed to improve the efficiency for genetic-based approaches in  searching neural networks from scratch for the future study.

% However, compared with other studies that aim to searching network under limited computational resources~\cite{liu2018darts,pham2018efficient}, our work has higher error rate and takes more running time. Although the studies in~\cite{liu2018darts,pham2018efficient} are conducted in different searching space compared with ours, the comparison still indicates  more exploration is needed to improve the efficiency for genetic-based approaches in  searching neural networks for the future study.

{\small
\bibliographystyle{ieee_full}
\bibliography{egbib}
}

\end{document}